\pdfoutput=1
\documentclass{article} %
\usepackage{nips15submit_e,times}
\usepackage{url}
\usepackage{amsmath}
\usepackage{amssymb}
\usepackage{bm}
\usepackage{pgf}
\usepackage{array}

\title{Learning to Learn Neural Networks}

\author{
Tom Bosc\\
INRIA, France\\
\texttt{tom.bosc@inria.fr} \\
}

\DeclareMathOperator*{\argmin}{\arg\!\min}

\nipsfinalcopy %
\usepackage[numbers]{natbib}
\usepackage{hyperref}

\begin{document}

$\newcommand{\Var}{\mathrm{Var}}$
$\newcommand{\Cov}{\mathrm{Cov}}$ 

\maketitle

\begin{abstract}
Meta-learning consists in learning learning algorithms. We use a Long Short Term Memory (LSTM) based network to learn to compute on-line updates of the parameters of another neural network. These parameters are stored in the cell state of the LSTM. Our framework allows to compare learned algorithms to hand-made algorithms within the traditional train and test methodology.
In an experiment, we learn a learning algorithm for a one-hidden layer Multi-Layer Perceptron (MLP) on non-linearly separable datasets. The learned algorithm is able to update parameters of both layers and generalise well on similar datasets.
\end{abstract}

\section{Introduction}
Meta-learning is the process of learning to fit parameters, that is, learning to learn. In this article, we focus on learning on-line learning algorithms : the data arrives sequentially and is shown only once. Thus, the meta-learning system is forced to update the parameters of the model after each observation. 

Hochreiter et al.~\cite{hochreiter2001learning} show that it is possible to train a Recurrent Neural Network (RNN) to learn the coefficients of quadratic functions under supervision using gradient descent. At each timestep, the RNN is given the inputs of the current timestep and the previous target. With a simple objective -- correctly predicting the targets at each timestep -- the RNN learns to improve its predictions with appropriate updates of its internal state. They use a LSTM~\cite{hochreiter1997long} with an additional hidden layer between the LSTM block and the output. %
 
In this work, we introduce a separator flag that split each sequence into a train and a test set. This enables us to compare learned algorithms to hand-made algorithms. In an experiment, a LSTM-based network is trained to learn a small one-hidden-layer MLP on a binary classification task. The weights of the MLP are stored in the LSTM memory state. We show that potentially complex and deep networks could be learned using this technique.
\section{Framework}
Consider $N$ datasets $\{(x^{(i)}_1,y^{(i)}_1),...,(x^{(i)}_{n_i},y^{(i)}_{n_i})\} \in D_{train}$ where $(x^{(i)}_t, y^{(i)}_t)$ are $n_i$ pairs of inputs and targets that are i.i.d. according to the $i$-th distribution. Similarly, we have $N'$ other datasets in $D_{test}$. %
We drop the index $(i)$ notation when it is not necessary.

We use a deterministic RNN that breaks down in two parts. The first part, the model, computes an estimate $o_t$ of the targets $y_t$ given the inputs $x_t$. It is parametrised by $\theta_t$, the parameter estimate at time $t$, which is the hidden state of the RNN at each timestep. 
\begin{equation}
o_t = f_{\theta_{t}}(x_t)
\end{equation}
$\theta_t$ is updated at every timestep by the second part of the network, the learner. The learner is an on-line updating mechanism parametrised by $\alpha$ and $\theta_1$, the initial state. It uses the inputs $x_t$, the targets $y_t$, the prediction of the model $o_t$ to compute the updates $\theta_{t+1}$. In addition, it uses a special input $1_{t < \tau}$ (indicator function) which role is explained below.
\begin{equation}
\label{eq:update}
\theta_{t+1} = g_{\alpha}(x_t, y_t * 1_{t < \tau}, 1_{t < \tau}, o_t)
\end{equation}

We call \textit{learning} the process of learning using the learner, i.e. learning $\theta_t$. \textit{Meta-learning} is the process of learning the learner, i.e. learning $\alpha$ and $\theta_{1}$, the initial parameter value. Note that the initial state of the RNN $\theta_1$, which corresponds to some sort of prior is optimised during the process of meta-learning.
The above update scheme does not impose that we show a target at each timestep, thanks to $1_{t < \tau}$. It is useful at test time, when we want to use the model for prediction without having the targets. Our datasets, from both $D_{train}$ and $D_{test}$, are themselves divided into training sets and test sets. This division is artificial when we generate datasets (for meta-learning or evaluating the learner), but is not when we are actually using real data. Let's call $\tau_i$ the indices for each dataset that separate the training set from the test set. $({x^{(i)}_t, y^{(i)}_t)}$ belongs to the $i$-th training set for each $t < \tau_i$ and to the $i$-th test set for each $t \geq \tau_i$. The RNN takes inputs sequences that are created using an arbitrary ordering of the samples, where all the training samples are shown first.~\footnote{It should be possible to interlace training samples with test samples and alternate training and predicting phases, but as we want to be able to evaluate the learned algorithm against other algorithms, we will not do this.} As we do not show any targets for samples in the test sets, we pass this information as a binary flag so that the learner does not misinterpret zeros as targets. This flag is $1_{t < \tau}$.

\label{sec:cost}
In this work, meta-learning is a byproduct of learning to correctly predict targets. There are no constraints on the learner other than producing a model that does good prediction, avoiding overfitting and underfitting. Evaluation of a learning algorithm should be done on a held-out test set. That is why the meta-learning objective is an average loss on the test sets of datasets in $D_{train}$:
\begin{equation}
\argmin\limits_{\alpha, \theta_{1}} \frac{1}{N} \sum\limits^N_{i = 1} \frac{1}{n_i - \tau_i + 1} (\sum\limits_{t = \tau_i}^{n_i} L(f_{\theta_t}(x^{(i)}_t), y^{(i)}_t))
\label{cost_1}
\end{equation}
Although, we introduce an alternative cost function used only for training purposes. It is the mean of the loss of all samples from all datasets in $D_{train}$:
\begin{equation}
\argmin\limits_{\alpha, \theta_{1}} \frac{1}{N} \sum\limits^N_{i = 1} \frac{1}{n_i} \sum\limits_{t = 1}^{n_i} L(f_{\theta_t}(x^{(i)}_t), y^{(i)}_t)
\label{cost_2}
\end{equation}

On the one hand, the cost \eqref{cost_1} prevents any kind of hard-wiring between the outputs and the targets that could happen if we decide to repeat inputs and targets during several timesteps. Indeed, learners that cheat by copying the target and pasting it at subsequent timesteps will fail to predict correctly on the test set. On the other hand, meta-learning is much harder with \eqref{cost_1} because there isn't any supervision during the learning phase. In our experiments, we use cost \eqref{cost_1} to evaluate learning algorithms against each other, but use \eqref{cost_2} in order to carry out meta-learning.

\section{Implementation}
The learner parameters $\alpha$ and $\theta_1$ are optimised by a learning algorithm and the model parameters $\theta_t$ are optimised by the learner. In this work, $f$ and $g$ should be differentiable because the errors at each timestep are backpropagated not only through the learner but also through the model. We use a LSTM-based network to store the parameters $\theta_t$ of the model. Thus, the learner only learn to update the weights, that is, to compute useful LSTM gate parameters. Our framework slightly differ from~\cite{hochreiter2001learning} in that we first compute the prediction and then update the weights.~\footnote{That is why the learner can receive $x_t$ and $y_t$ at the same timestep and does not hard-wire the target to the output $o_t$. This does not make any difference in performance and is not an improvement over~\cite{hochreiter2001learning}.}

$g_{\alpha}$ is an LSTM block which gates are computed after the concatenation of the inputs went through several regular layers. It has $n$ fully connected layers of the form : $g_i(x) = h_i(W_i x + b_i)$ with $h_i$ an activation function such as tanh, sigmoid, Rectified Linear Unit (ReLU), etc. The output of the last layer $x^*$ is used to compute the gate parameters of the LSTM block. There isn't any LSTM output gate because the output at each timestep is provided by the model.
\begin{equation}
\begin{split}
&x^* = g_1 \circ ... \circ g_n ([x_t, y_t * 1_{t < \tau}, 1_{t < \tau}, o_t]) \\
&z_t = W_z x^* + b_z \\
&i_t = \sigma(W_i x^* + b_i) \\
&f_t = \sigma(W_f x^* + b_f) \\
\end{split}
\end{equation}
The parameters at each timestep are given by :
\begin{equation}
\theta_{t+1} = g_{\alpha}(x_t, y_t * 1_{t < \tau}, 1_{t < \tau}, o_t) = i_t * z_t + f_t * \theta_{t}
\end{equation}

It should be noted that the learner and the model are inseparable. Each component of each gate parameter in the learner is bound to a specific parameter in the model. That is also why the learner can learn without seeing the parameters of the model as an input. But it could be argued that the behavior of certain parameters of the model (for example, units of the same layer in the case of a neural network) could and should be generalised. This criticism could be addressed in a future work.
\section{Experiment}
We train the learner described above to fit parameters of a neural network on binary classification tasks. The model is a one-hidden layer MLP of $n_{in} = 5$ inputs, 32 hidden units and 1 output. The activation function of the first layer is a ReLU and the second is the sigmoid function. Counting the biases, there are $225$ parameters to learn. The learner contains two layers of size $128$ and $256$ before the LSTM gates that are activated by ReLUs. There are $207651$ meta-learnable parameters. The loss function is cross-entropy.

Our training and test datasets are generated randomly in a hierarchical way. The process described here is arbitrarily designed and creates noisy non-linearly separable data. We generate one random covariance matrix per dataset by computing the product of a $n_{in} \times n_{in}$ matrix of random coefficients (uniformly distributed from -1 to 1) with its transpose. It is the covariance of a multivariate Gaussian that produces samples X for this dataset. In order to model noise, we draw from an exponential distribution (of parameter $\beta=2$) a parameter $\epsilon_i$ for each dataset. It is then used to draw a vector $v_i$ of size $n_{in}$ from the exponential distribution of parameter $\beta=\epsilon_i$. Noise is then generated using a multivariate Gaussian of mean 0 and covariance defined by the diagonal matrix of diagonal $v_i$, and added to the samples. Then, we partition our inputs twice by applying a randomly chosen linear transformation and thresholding it to create binary classes. Then, a logical XOR is applied to obtain the targets. We throw away datasets which classes balances are below a minimal threshold (0.15). Postprocessing of the inputs include centering and scaling to unit variance. 

The type and quantity of noise can be seen as a prior on the data. But it may also ease the optimisation process by introducing noise in the gradients, hypothesis which remains to be investigated.

We use a gradient descent algorithm called SMORMS3\cite{funk2015rmsprop} with a learning rate of $10^{-3}$ on one randomly drawn sequence from $D_{train}$ per iteration (Stochastic Gradient Descent). 

\begin{figure}[h]
\begin{center}
\scalebox{0.33}{\includegraphics{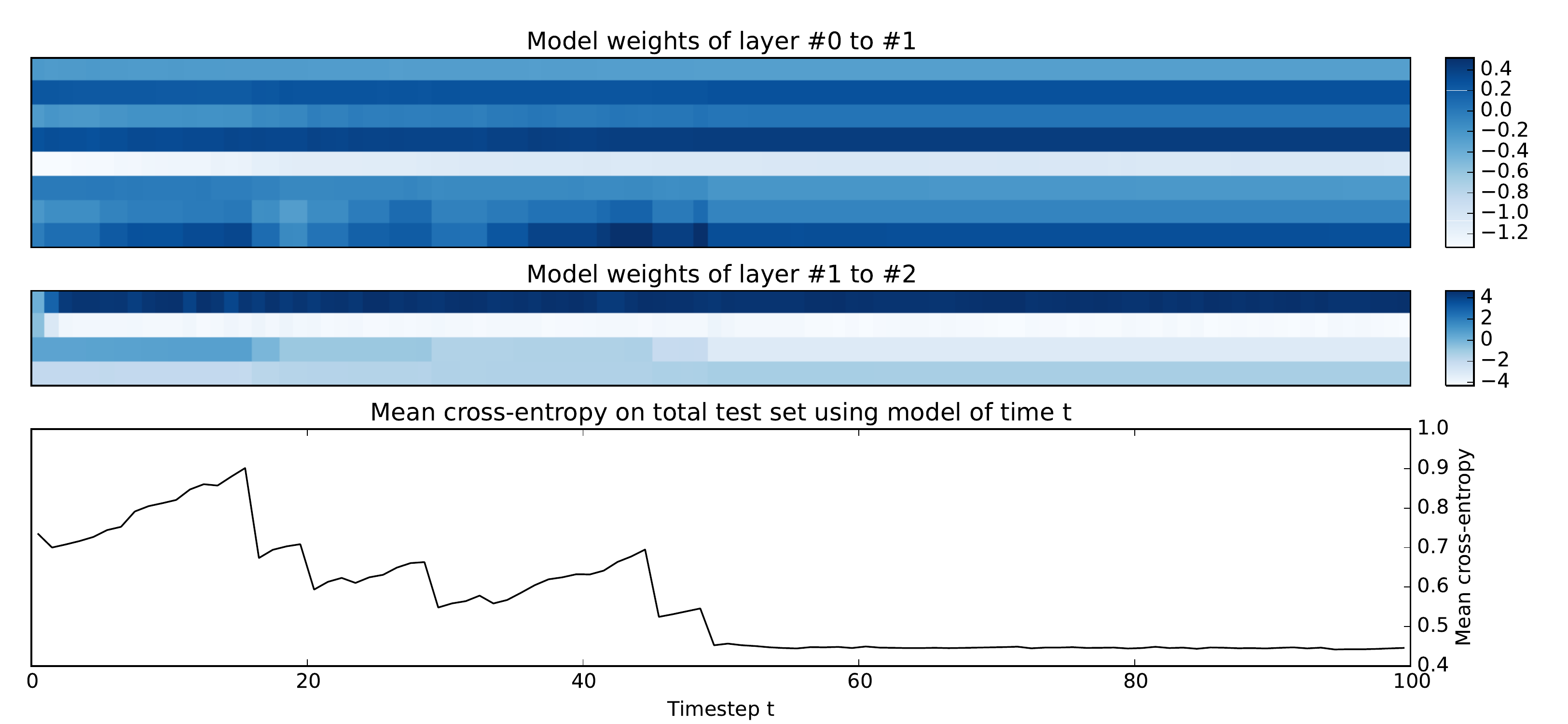}}
\end{center}
\caption{Illustration of the learning process on an example dataset split in training and test sets of the same size (50 samples each). The input space is $\mathbb{R}^{2}$ and the model is smaller than the one described above (2 inputs, 4 hidden units, 1 output).
The matrices show the weights of the model across time. The biases are not shown. The plot at the bottom shows the mean cross-entropy on the entire test set using the model $\theta_t$.}
\label{model_1}
\end{figure}

In order to test whether the non-linearity of the model is effectively used by the learner, we present in the following array the mean and standard deviation of mean-cross entropy losses of hand-made algorithms against a learned algorithm. $D_{test}$ consists of 500 randomly generated datasets of 200 samples each, with variable train over test ratios. The noise parameter $\beta$ is lowered to $1$ so that there is less noise than during meta-learning. Logistic regression is regularised with either L1 or L2, with $\lambda \in \{0.1, 1.0, 10\}$ and the hyperparameters are chosen with k-fold cross-validation. %
\begin{small}
\begin{center}
\begin{tabular}{|l|l|l|l|l|}
\hline 
& LSTM & Logistic Regression & SVM (linear) & SVM (RBF)  \\
\hline 
$\mu$ MCE & 0.540	& 0.574 & 0.573 & 0.507 \\
\hline 
$\sigma$ MCE & 0.139	& 0.208 & 0.159 & 0.164 \\
\hline
\end{tabular}
\end{center}
\end{small}
The learned algorithm achieves a significantly lower cost on average than the linear models. It generalises on non-separable data, which we know is the limit of these models. Although, it is worse than SVM with a RBF kernel. A proper evaluation that compare performance of the same MLP model trained with different learning algorithms is needed and is left for future work.

Figure~\ref{model_1} shows details of the learning process. The weight matrices indicates that some parameters are modified by the learner while others are left aside and practically do not change over time. This illustrates a trade-off between learning and meta-learning. One explanation is that the expressivity of the learner does not allow it to compute good updates of all the weights. It is better to let the meta-learning process select a subset of parameters which will will not be updated, optimise their initial value and focus on learning to update other parameters which provide relatively easy performance gains. Another possible explanation is that the optimisation process was stopped too early or that $D_{train}$ contains too few datasets.%

The parameters that are updated by the learner and the parameters updated during meta-learning are scattered across the two weight matrix of the model. It indicates that the learner can compute updates for neural networks that are more complicated than shallow networks. 

The first half of the last plot can be seen as a learning curve. %
The loss is aligned with the updates made by the learner so the updates effectively improves the loss on the test set. Although individually, not all updates improve the performance on the test set. After having seen all the train samples, at iteration 50, the performance stops changing. The second halves of the weight matrices indicate that the learner stops to update the weights of the model on the test sets. With our architecture, nothing forbids the learner to keep modifying the parameters. A possible direction of research is to encourage updates during the test phase so that the learner simulate sampling from a distribution of parameters instead of using the same point-estimate over the whole test set.

\section{Perspectives and conclusion}
It is possible to train a neural network to train another neural network using backpropagation. More work is needed to characterise both the learning and the meta-learning process, in terms of generalisation and regularisation abilities as well as in speed of convergence. 
\bibliographystyle{unsrt}  
\bibliography{references}
\end{document}